\documentclass{ELSP}
\PassOptionsToPackage{unicode}{hyperref}
\PassOptionsToPackage{hyphens}{url}
\IfFileExists{upquote.sty}{\usepackage{upquote}}{}
\IfFileExists{microtype.sty}{% use microtype if available
	\usepackage[]{microtype}
	\UseMicrotypeSet[protrusion]{basicmath} % disable protrusion for tt fonts
}{}
\makeatletter
\IfFileExists{parskip.sty}{%
	\usepackage{parskip}
}
\makeatother
\usepackage{soul}
\usepackage{xcolor}
\usepackage{bm}
\IfFileExists{xurl.sty}{\usepackage{xurl}}{} % add URL line breaks if available
\IfFileExists{bookmark.sty}{\usepackage{bookmark}}{\usepackage{hyperref}}
\hypersetup{
	pdfcreator={ELSP},
	pdftitle={A high-fidelity digital twin for robotic manipulation based on 3D Gaussian Splatting},
	pdfauthor={Ziyang Sun and Lingfan Bao and Tianhu Peng and Jingcheng Sun and Chengxu Zhou},
	pdfsubject={Robotics; 3D Gaussian Splatting; digital twin; sim-to-real transfer},
	pdfkeywords={3D Gaussian Splatting; digital twin; robotic manipulation; Real-to-Sim-to-Real}}
\usepackage{longtable,booktabs,array}
\usepackage{calc} % for calculating minipage widths% for algorithm environment
\usepackage{algorithmic}
\usepackage{etoolbox}
\makeatletter
\patchcmd\longtable{\par}{\if@noskipsec\mbox{}\fi\par}{}{}
\makeatother
\IfFileExists{footnotehyper.sty}{\usepackage{footnotehyper}}{\usepackage{footnote}}
\makesavenoteenv{longtable}
\usepackage{setspace}
\usepackage{graphicx}
\usepackage{amsmath}
\usepackage{times}
\usepackage{mathptmx}
\usepackage{caption}
\usepackage{subcaption}
\usepackage{geometry}
\usepackage{doi}

\usepackage{float}
\usepackage{enumerate}
\usepackage{enumitem}
\usepackage{stfloats}
\usepackage{ragged2e}
\usepackage{titlesec}
\usepackage[none]{hyphenat}
\usepackage[numbers,sort,compress]{natbib}
\setcitestyle{numbers,square,comma,sort&compress}
\newlength\mylen
\setcitestyle{citesep={,\kern-\mylen}}

\hypersetup{
    colorlinks=false, 
    pdfborder={0 0 0} 
}
\sethlcolor{yellow}

\usepackage{algorithm}

\titleformat{\section}[hang]
  {\fontsize{12pt}{12pt}\selectfont\bfseries\color[RGB]{0,131,255}} 
  {\thesection}{0.5em}{}

\titleformat{\subsection}[hang]
  {\fontsize{12pt}{12pt}\selectfont\itshape} 
  {\thesubsection}{0.5em}{}

\titleformat{\subsubsection}[hang]
  {\fontsize{12pt}{12pt}\selectfont} 
  {\thesubsubsection}{0.5em}{}

\usepackage{fancyhdr}
\fancypagestyle{firstpage}
{
	\fancyhf{}
	\setlength{\headsep}{6px}
	\fancyhead[R]{\sffamily \fontsize{9pt}{8pt}\selectfont \textbf{\textcolor[RGB]{0,131,255}{Robot Learn.}}}
	\fancyhead[L]{\sffamily \fontsize{9pt}{8pt}\selectfont \textbf{\textcolor[RGB]{0,131,255}{ELSP}}}
	\fancyfoot[L]{\sffamily \fontsize{9pt}{8pt}\selectfont Sun Z, {\em et al}. Robot Learn. 2026(2):000x}
}

\begin{document}

\thispagestyle{firstpage}

\let\thefootnote\relax
\footnotetext{
\newline
\vspace{12pt}
    \raisebox{-\height}[0pt][0pt]{
        \begin{minipage}[h]{\linewidth}
            \begin{minipage}[h]{0.15\linewidth}
                \includegraphics[width=\linewidth, height=1cm]{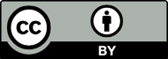}
            \end{minipage}
            \hfill
            \begin{minipage}[h]{0.82\linewidth}
            \justifying
                \footnotesize Copyright©2026 by the authors. Published by ELSP. 
                This work is licensed under a Creative Commons Attribution 4.0 
                International License, which permits unrestricted use, distribution, 
                and reproduction in any medium provided the original work is properly cited.
            \end{minipage}
        \end{minipage}
    }
}

\begin{spacing}{0.88}
{\sffamily \small \noindent {\textls[-55]{Article $\mid$ Received 11 December 2025; Revised 22 February 2026; Accepted 6 April 2026; Published XX April 2026}}}\\
{\sffamily\small{https://doi.org/10.55092/rl2026000x}}
\end{spacing}

\setstretch{1.24}

% \vspace{10pt}
% {\raggedright
% \papertitle{%
% \makebox[\textwidth][s]{A high-fidelity digital twin for robotic manipulation based%
%   \hspace*{\fill}%
%   \href{https://crossmark.crossref.org/dialog/?doi=10.55092/rl2026000x}{%
%     \raisebox{-1.2ex}{\includegraphics[width=0.93cm,height=0.93cm]{figure/Check for updates.png}}%
%  }%
% }%
% \\[-0.2ex]
% on 3D Gaussian Splatting%
% }%
% }

\vspace{10pt}
{\raggedright
\papertitle{%
A high-fidelity digital twin for robotic manipulation based on 3D Gaussian Splatting%
}%
}

\vspace{12pt}
\hspace{-0.91cm}\authorname{Ziyang} {Sun} {}\textbf{,}
 \hspace*{-0.1cm}\authorname{Lingfan} {Bao} {}\textbf{,}
  \hspace*{-0.1cm}\authorname{Tianhu} {Peng} {}\textbf{,}
  \hspace*{-0.1cm}\authorname{Jingcheng} {Sun} {} 
\textbf{and}
 \hspace*{-0.1cm}\authornameCorres{Chengxu} {Zhou}{}{*}

\vspace{12pt}
\hspace{-1.06cm}\formatintroduction{}{Department of Computer Science, University College London, London, UK}

\vspace{-10pt}
\authoremail{Correspondence author} {chengxu.zhou@ucl.ac.uk.}

\vspace{12pt}
\noindent\textbf{\textcolor[RGB]{0,131,255}{Highlights:}}
\vspace{12pt}
\begin{itemize}[left=0pt, labelwidth=0pt, labelsep=17pt, itemsep=0pt]
\setlength{\itemsep}{0pt} 
\setlength{\parskip}{0pt} 
    \item Proposes a unified perception-to-planning-to-execution workflow leveraging 3DGS to create photorealistic digital twins from sparse RGB views in minutes, forming a holistic, closed-loop pipeline from capture to real-robot execution.
    
    \item Introduces a robust method for semantic understanding by lifting 2D masks from foundation models like SAM into the 3D scene using multi-view spatial consensus, and an efficient conversion of raw 3DGS into planning-ready collision geometry for physics-based simulation.
    
    \item Demonstrates real-world effectiveness with physical experiments on a Franka Emika Panda robot, showing that motion plans validated in the digital twin can be transferred to real-robot execution.
\end{itemize}

\vspace{10pt}
\noindent\textbf{\textbf{\textcolor[RGB]{0,131,255}{Abstract:}}} 
\textls[+4]{Developing high-fidelity, interactive digital twins is crucial for enabling closed-loop motion planning and reliable real-world robot execution, which are essential to advancing sim-to-real transfer. However, existing approaches often suffer from slow reconstruction, limited visual fidelity, and difficulties in converting photorealistic models into planning-ready collision geometry. We present a practical framework that constructs high-quality digital twins within minutes from sparse RGB inputs. Our system employs 3D Gaussian Splatting (3DGS) for fast, photorealistic reconstruction as a unified scene representation. We enhance 3DGS with visibility-aware semantic fusion for accurate 3D labelling and introduce an efficient, filter-based geometry conversion method to produce collision-ready models seamlessly integrated with a Unity-ROS2-MoveIt physics engine. In experiments with a Franka Emika Panda robot performing pick-and-place tasks, we demonstrate that this enhanced geometric accuracy effectively supports robust manipulation in real-world trials. These results demonstrate that 3DGS-based digital twins, enriched with semantic and geometric consistency, offer a fast, reliable, and scalable path from perception to manipulation in unstructured environments.}

\vspace{12pt}
\noindent\textbf{\textcolor[RGB]{0,131,255}{Keywords:}} 3D Gaussian Splatting; digital twin; robotic manipulation; Real-to-Sim-to-Real

\section{Introduction}

\textls[+10]{The field of robotics is rapidly moving towards full autonomy in complex and unstructured environments. Effective autonomous manipulation in unstructured environments fundamentally relies on the robot's ability to rapidly construct a high-fidelity, actionable understanding of its surroundings, which is a core requirement for achieving advanced tasks such as fine-grained manipulation. This actionable understanding relies heavily on the construction of an accurate virtual replica, commonly known as a digital twin~\cite{turn0search4}.  Digital twins are essential tools that enable safe, repeatable validation, closed-loop motion planning, and reliable sim-to-real transfer, which is important for advancing the deployment of robotic systems in the real world.}

However, existing reconstruction pipelines~\cite{turn0search1} introduce significant bottlenecks that impede their seamless integration into real-time robotic workflows. The pursuit of visual fidelity often conflicts with the need for computational efficiency and physical utility. Specifically, Neural Radiance Fields (NeRF)~\cite{mildenhall2020nerfrepresentingscenesneural,barron2021mipnerfmultiscalerepresentationantialiasing}  offer high photorealism but are computationally expensive, often requiring minutes to hours of time for optimisation, which severely limits rapid deployment. Conversely, traditional methods based on point clouds or mesh reconstruction~\cite{Lv_2022} are faster but often suffer from insufficient fidelity, noise sensitivity, and the difficulty of projecting reliable, consistent semantic labels from sparse multi-view inputs.

\textls[-8]{In this context, 3DGS~\cite{kerbl20233dgaussiansplattingrealtime}, a novel explicit radiance field reconstruction method, has shown outstanding performance in both reconstruction quality and speed and has emerged as a potential breakthrough representation. 3DGS successfully balances the trade-off between speed and fidelity, achieving photorealistic rendering quality within minutes, which is highly promising for rapid robotic scene capture.} However, the core representation used by 3DGS, anisotropic Gaussian splats (or ``balls''), defines each scene primitive not merely by a point but by a position, a covariance matrix, a colour, and an opacity. Although this representation excels at combining colour and alpha (opacity) components to create a visually convincing and continuous surface view, the underlying explicit geometry remains inherently ambiguous and problematic for physical interaction. Specifically, the resulting cloud of stretched Gaussian primitives is not a clean, watertight surface, but is riddled with reconstruction artefacts. These issues include floaters (isolated clusters derived from optimisation residuals), ghost artefacts (semi-transparent, low-opacity points near reflective or occluded boundaries), and overall surface fuzziness. These geometric imperfections, while visually hidden in the rendered image, render the raw 3DGS output unsuitable for precise robotic tasks such as collision checking and motion planning.

The fragmentation between these approaches highlights a critical, unsolved challenge: generating a digital twin that is simultaneously photorealistic, rapidly reconstructed, and equipped with planning-ready collision geometry and consistent semantic structures. Addressing this triple constraint, speed, fidelity, and actionability, is the central challenge for next-generation robotic perception systems.

To bridge the gap between visual fidelity, reconstruction efficiency, and physical utility in digital twin construction, we present a unified framework that generates interactive, semantically structured digital twins directly from sparse RGB inputs. 

To ensure the resulting environment is geometrically interactive and planning-ready, we introduce two critical components: 
(i) a visibility-aware semantic fusion module that aggregates multi-view cues~\cite{kirillov2023segment, ren2024groundedsamassemblingopenworld} to achieve consistent 3D semantic labelling of the Gaussian primitives, and 
(ii) a geometric refinement process that addresses the inherent limitations of Gaussian-based representations—specifically the presence of floater artefacts and noisy density distributions—by converting these primitives into precise, collision-ready meshes. This conversion is vital for transforming a purely visual reconstruction into a physically actionable asset that meets the strict geometric requirements of motion planners.

While recent initiatives like Splat-Nav~\cite{chen2025splatnav} and Splat-MOVER~\cite{shorinwa2024splatmover} have integrated 3DGS into robotic workflows, they predominantly target mobile navigation or semantic affordance detection, often relying on coarse occupancy proxies that lack the geometric fidelity required for fine-grained manipulation. Similarly, RoboGSim~\cite{li2025robogsimreal2sim2realroboticgaussian} represents a significant step forward by utilising 3DGS to construct photorealistic environments for offline reinforcement learning. However, its primary focus lies in bridging the visual domain gap for policy training rather than enabling rapid, online geometric perception. In contrast, our framework prioritises immediate physical actionability by focusing on rapid planning-ready reconstruction, converting image-level input into collision-ready meshes within minutes to support precise, solver-based motion planning in the real world.

\textls[-15]{Finally, we deploy these assets into a custom Unity-based environment to facilitate a validated Real-to-Sim-to-Real workflow, enabling motion plans generated within the high-fidelity digital twin to be validated before transfer to physical robot execution. This closed-loop pipeline, integrating photorealistic reconstruction, semantic understanding, collision geometry, and validated simulation execution, represents progress toward fully actionable digital twins for complex robotic manipulation.}

The main contributions of this work are summarised as follows:
\begin{enumerate}[label=(\arabic*)]\setlength{\leftskip}{+0.48cm}

    \item Unified Real-to-Sim-to-Real Framework:
    \textls[+10]{We propose a unified Real-to-Sim-to-Real framework that synergises 3DGS with robust point cloud processing. This approach generates actionable digital twins within minutes, effectively bridging the gap between neural rendering and \mbox{robotic manipulation}.}    
    \item Visibility-Aware Semantic Fusion: We introduce a view-dependent semantic aggregation with occlusion-aware confidence weighting strategy. This method distils 2D segmentation cues from vision foundation models~\cite{kirillov2023segment,ravi2024sam2segmentimages} into consistent 3D attributes, resolving projection ambiguities to ensure accurate labelling under occlusion and providing a semantically structured environment for robot manipulation.
    \item Planning-Ready Geometry Conversion: \textls[-20]{Addressing the limitation of 3DGS as a purely visual representation, we implement a multi-stage geometric refinement process with attribute-based pruning, connectivity analysis, and alpha-shape meshing. By combining opacity-based thresholding, scale-based filtering, and DBSCAN clustering before mesh generation, this step converts raw Gaussian splat point clouds into planning-ready collision meshes.}
    \item Real-World Closed-Loop Validation: Beyond simulation-only validation, we demonstrate a complete perception-planning-validation-execution loop on a Franka Emika robot~\cite{coleman2014reducingbarrierentrycomplex}, showing that plans generated from the converted collision geometry transfer to physical execution.
\end{enumerate}

\textls[-18]{The remainder of this paper is organised as follows. Section~2 reviews related work on 3D reconstruction, semantic understanding, and digital twins for manipulation. Section~3 presents the proposed methodology, including high-fidelity reconstruction, visibility-aware semantic fusion, physics-ready geometry reconstruction, and system integration. Section~4 describes the experimental setup, evaluation metrics, and experimental results. Section~5 discusses the limitations of the current framework. Section~6 concludes the paper and outlines future work.}

\section{Related work}

\subsection{3D scene reconstruction for robotics}

\textls[-12]{While dense mapping pipelines like TSDF~\cite{curless1996volumetric} and Voxblox~\cite{oleynikova2017voxblox} have long served as the backbone for robotic navigation, their dependence on voxel discretisation fundamentally limits their utility for manipulation. The resulting over-smoothed geometry fails to capture the high-frequency surface details necessary for fine motor control. Similarly, while implicit representations such as NeRF~\cite{mildenhall2020nerfrepresentingscenesneural} and Instant-NGP~\cite{M_ller_2022} offer visual photorealism, they remain constrained by prohibitive inference latencies and dense view requirements, creating a bottleneck for real-time robotic exploration.}

Recently, 3DGS~\cite{kerbl20233dgaussiansplattingrealtime} has emerged to bridge this gap by explicitly representing scenes as anisotropic Gaussian primitives, combining differentiable optimisation with fast rasterisation, offering photorealistic rendering at real-time speeds. \textls[-12]{However, despite this visual fidelity, raw 3DGS representations are inherently ill-suited for physical interaction.} The presence of reconstruction artefacts, such as floaters, ghost points, and surface fuzziness, makes the resulting point clouds functionally unsuitable for direct collision checking and robot manipulation.

In contrast, our framework is specifically designed to close the gap between visual realism and physical validity. By systematically structuring raw 3DGS outputs, we transform noisy visual primitives into manipulation-ready geometry without sacrificing the rendering speed required for online operation.

\vspace{-4pt}
\subsection{Semantic scene understanding}

Robotic manipulation requires a precise understanding of object identity beyond mere geometry. Although foundation models like SAM~\cite{kirillov2023segment} and Grounded SAM~\cite{ren2024groundedsamassemblingopenworld} have revolutionised 2D segmentation, lifting these predictions into 3D space remains a critical challenge. Existing feature distillation methods, such as SegmentAnyGaussian~\cite{cen2025segment3dgaussians}, attempt to solve this by appending high-dimensional vectors to primitives, but this approach drastically increases memory footprints and training overhead, limiting deployment agility. Conversely, direct projection methods often suffer from ``bleeding'' labels and inconsistencies caused by depth discontinuities.

What sets our approach apart is that it diverges from these computationally heavy or unstable methods by introducing a Visibility-Aware Semantic Fusion module. Instead of relying on extensive retraining or naive projection, our pipeline is grounded in a rigorous geometric consensus mechanism. This method integrates depth and visibility checks with a confidence-weighted voting scheme, ensuring that semantic labels are not just projected, but geometrically verified across views. This foundation enables our system to achieve high-fidelity 3D labelling that is both consistent and computationally lightweight.

\vspace{-4pt}
\subsection{Digital twins for interactive manipulation}

The ultimate goal of robotic perception is to enable interaction. Ideally, a digital twin must unify three capabilities: photorealistic rendering, semantic understanding, and physical collision handling. Traditional simulators (e.g., Gazebo, MuJoCo) achieve physics but lack visual fidelity, while neural renderers achieve fidelity but lack physical structure. Bridging this ``Sim-to-Real'' gap requires a hybrid representation.

\textls[+12]{Recent efforts have begun to integrate 3DGS into planning frameworks. Splat-Nav \cite{chen2025splatnav} utilises Gaussian representations for navigation, employing ellipsoid abstractions for collision checking. However, its focus is global path planning rather than the object-level granularity required for grasping.} Similarly, \mbox{RoboGSim \cite{li2025robogsimreal2sim2realroboticgaussian}} focuses on the simulation aspect, providing a platform for offline testing rather than an online perception-to-action pipeline. Other works like Splat-MOVER~\cite{shorinwa2024splatmover} and GraspSplats~\cite{ji2024graspsplats} explore open-vocabulary manipulation and 3D feature splatting for grasping, respectively.

Despite these advancements, existing frameworks face fundamental gaps regarding manipulation. Firstly, the presented works do not provide robust mechanisms to convert raw, noisy 3DGS outputs (often plagued by floaters and ghost artefacts) into planning-ready collision geometry. Secondly, existing works primarily address visual policy learning or navigation, leaving the validation of classical motion planning on reconstructed geometry unexplored. In particular, RoboGSim~\cite{li2025robogsimreal2sim2realroboticgaussian} provides a valuable 3DGS-enabled simulation platform for robot learning. However, it primarily utilises 3DGS for photorealistic rendering rather than as a source for collision geometry generation. The systematic conversion of Gaussian primitives into planning-compatible collision geometry, and its validation through real-robot execution, remain underexplored. Our work directly addresses both challenges. Addressing these limitations requires a unified pipeline that integrates sparse-view reconstruction~\cite{fan2025instantsplatsparseviewgaussiansplatting}, geometrically-verified semantic consensus, and physics-based post-processing~\cite{kazhdan2006poisson,guedon2023sugar} into a reliable robotic workflow.

\begin{figure}[H]
    \centering
    \includegraphics[width=\textwidth]{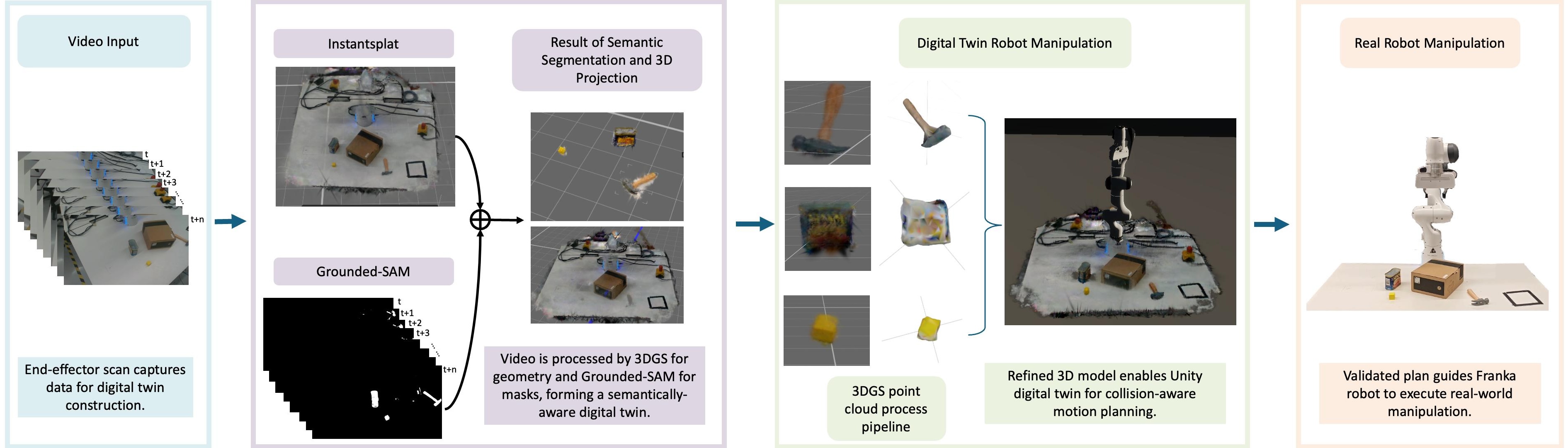}   
    \caption{The overall pipeline of this framework uses multi-view video input and 3DGS to reconstruct the scene geometry. Grounded SAM provides semantic masks, which are fused with the 3D projection to form a semantically aware digital twin. This twin enables collision-aware motion planning for real robot manipulation.}
    \label{fig:pipeline}
\end{figure}

\textls[+12]{Our proposed framework addresses this fragmentation by establishing a unified closed-loop Real-to-Sim-to-Real pipeline, as shown in Figure~\ref{fig:pipeline}. It seamlessly integrates photorealistic 3DGS reconstruction with our refined collision geometry (using alpha-shape meshing), semantic understanding, and a Unity-based robot simulation and manipulation interface~\cite{unitygaussiansplat,ros2forunity}.} This complete, validated sim-to-real workflow is a key advancement, ensuring that motion plans generated against the high-fidelity digital twin are reliable for real-world execution on the real robot.
\vspace{+10pt}

\section{Methodology}

Our proposed framework establishes a comprehensive digital twin generation pipeline tailored for closed-loop robotic manipulation. Following the framework pipeline illustrated in Figure~\ref{fig:pipeline}, the system operates via two parallel processing streams: (1) a geometric reconstruction stream that uses an optimised 3DGS approach to generate a high-fidelity 3D scene, and (2) a semantic segmentation stream that identifies and isolates manipulable objects. The subsequent stages focus on rigorously transforming this semantically-annotated 3D model into clean, planning-ready collision geometry and integrating it into the simulation environment for validation.

\vspace{-4pt}
\subsection{High-fidelity scene reconstruction}

\textls[-10]{We employ a 3DGS-based approach for scene reconstruction due to its superior balance of rendering quality and rapid optimisation speed. This choice is important for achieving fast reconstruction and high-quality digital twins, contrasting sharply with NeRF-based methods whose latest algorithms commonly require tens of minutes or more for comparable fidelity. To address the challenges posed by sparse and uncalibrated input images, which often lead to failures in traditional Structure-from-Motion (SfM) pipelines,} we use the InstantSplat methodology~\cite{fan2025instantsplatsparseviewgaussiansplatting}. This streamlined approach eliminates the need for a separate SfM step by utilising a pre-trained geometric prior, such as MASt3R~\cite{leroy2024groundingimagematching3d}, to directly estimate an initial point cloud and camera poses.

The core of the reconstruction is a self-supervised optimisation process. The scene is represented by a set of 3D Gaussians, each defined by a position $\bm{\mu}$, a covariance matrix $\bm{\Sigma}$, a colour, and an opacity. The unnormalised density is given by~\cite{kerbl20233dgaussiansplattingrealtime}:
\vspace{-8pt}
\begin{equation}
G(\bm{x}) = \exp\left( -\frac{1}{2} (\bm{x}-\bm{\mu})^\top \bm{\Sigma}^{-1} (\bm{x}-\bm{\mu}) \right)
\end{equation}

\vspace{-4pt}
This set of Gaussians is jointly optimised with the camera poses to minimise the photometric rendering error between the rendered images and the input views. This technique bypasses the traditional, time-consuming adaptive density control steps of vanilla 3DGS, enabling extremely fast convergence and yielding a high-fidelity 3D representation suitable for photorealistic rendering and depth extraction.

\vspace{-4pt}
\subsection{Visibility-aware semantic fusion}
\label{subsec:semantic_lifting}

Achieving a reliable, actionable semantic understanding is the prerequisite for robot interaction. However, a fundamental conflict arises when lifting 2D perception to 3D: standard single-view segmentation models (e.g., SAM) suffer from the ``bleeding effect,'' where background pixels near the object boundary are erroneously included in the foreground mask. To resolve this, we propose a visibility-aware semantic fusion framework. Unlike naive projection methods, our approach treats 2D masks as noisy spatial hypotheses and enforces 3D geometric consistency to filter out segmentation outliers.

Spatial Isolation via Depth Clustering:
The core premise of our method is that semantic coherence implies spatial coherence. While a 2D segmentation mask $M_j$ may loosely cover both the target object and the adjacent background, the underlying 3D geometry exhibits a distinct depth discontinuity. 

To exploit this, we perform Depth-Guided Isolation for each view. We project the 3D Gaussians contained within the 2D mask and apply density-based clustering (DBSCAN) on their depth values. We assume the largest cluster corresponds to the true object geometry, while smaller, spatially detached clusters represent background artefacts included by the 2D model. 

Confidence-Weighted Consensus:
To aggregate these observations into a unified 3D semantic field, we employ a weighted voting mechanism governed exclusively by spatial validity. We define the fusion weight $W_{i,j}$ for Gaussian $i$ in view $j$ as:
\vspace{-12pt}
\begin{equation}
W_{i,j} = w_{\text{cluster}}(i, j)
\end{equation}

\vspace{-6pt}
Here, $w_{\text{cluster}}$ acts as a \textit{soft spatial gate}. Points belonging to the primary depth cluster are assigned high confidence ($w_{\text{cluster}} \approx 1$), while spatial outliers are suppressed ($w_{\text{cluster}} \approx 0$). This formulation is robust against the geometric ambiguities of the 2D masks, ensuring that only points physically co-located with the object contribute to the semantic label.

The final semantic label for a 3D point $\mathbf{p}_{i}$ is determined by accumulating these spatially-weighted votes across all visible views $V'_i$:

\begin{equation}
\mathbf{p}_{i} \in
\begin{cases}
\mathcal{P}_{\text{obj}} & \text{if } \sum_{j \in V'_i} W_{i,j} \cdot \mathbb{I}((u_i, v_i) \in M_j) \geq \tau_{\text{consensus}} \\
\mathcal{P}_{\text{bg}} & \text{otherwise}
\end{cases}
\end{equation}

\vspace{+2pt}
Notation: $\mathbf{p}_i$ denotes the 3D position of Gaussian $i$ (or its mean $\bm{\mu}_i$). $(u_i,v_i)$ denotes the 2D projection of $\mathbf{p}_i$ into view $j$ under the estimated camera model. $M_j$ is the 2D foreground mask in view $j$. $\mathbb{I}(\cdot)$ is an indicator function that equals 1 if the predicate is true. $V'_i$ is the set of views where Gaussian $i$ is visible after the depth/visibility test. $W_{i,j}$ is the spatial validity weight defined in Equation~(2). $\tau_{\text{consensus}}$ is the consensus threshold controlling the completeness/artefact trade-off.

Table~\ref{tab:notation_semantic} summarises the main notations used in the semantic lifting formulation. These symbols define the 3D point representation, image projection, mask observation, visibility filtering, spatial weighting, and multi-view consensus threshold used in our semantic reconstruction pipeline.
\begin{table}[H]
\centering
\fontsize{10}{12}\selectfont
\caption{Main notations used in semantic lifting.}
\label{tab:notation_semantic}
\begin{tabular}{ll}
\toprule
\textbf{Symbol} & \textbf{Meaning}  \\
\midrule
$\mathbf{p}_i$ & 3D point (Gaussian mean) of primitive $i$ \\
$(u_i,v_i)$ & Pixel coordinate of $\mathbf{p}_i$ projected to view $j$ \\
$M_j$ & 2D segmentation mask in view $j$ \\
$V'_i$ & Visible-view set for primitive $i$ after depth/visibility checks \\
$W_{i,j}$ & Spatial validity weight for primitive $i$ in view $j$ \\
$\tau_{\text{consensus}}$ & Multi-view vote threshold \\
$N$ & Number of visible views, $N=|V'_i|$ \\
$\kappa$ & Strictness factor for the consensus threshold \\
\bottomrule
\end{tabular}
\end{table}

\vspace{-4pt}
\noindent In practice, we set $\tau_{\text{consensus}} = N/\kappa$. This setting makes the vote requirement increase with view coverage while remaining tolerant to occasional segmentation failures. Based on the consistency-artefact trade-off analysed in Section~4.4.1, we use $\kappa=1.5$ as the default value.

For DBSCAN-based depth clustering, we empirically determine hyper-parameters to balance object connectivity and artefact rejection. We set $\varepsilon_{\text{depth}} = 0.02\,\text{m}$, which corresponds to approximately 2\% to 3\% of the typical workspace depth range (0.3 to 1.0\,m). This threshold effectively groups Gaussian primitives belonging to the same physical surface (typically < 2\,cm apart in depth) while rejecting background floaters and ghosting artefacts that are spatially detached (> 5\,cm separation). The minimum cluster size is set to min\_samples = 10 to filter out spurious small clusters arising from segmentation noise, a value validated across our object set ranging from 5 to 30\,cm in size. These parameters proved robust across all test scenarios. However, deployment at substantially different scales (e.g., warehouse logistics or micro-manipulation) may require recalibration, as discussed in the Limitations section.

\textls[-25]{This consensus mechanism effectively ``carves'' the correct semantic shape out of the noisy 2D predictions.}

\textls[-12]{Iterative Semantic Refinement:
To further sharpen the boundaries, we implement an iterative feedback loop ($\text{max\_iter}=3$). As the 3D semantic model improves, it generates cleaner depth maps for the next iteration's visibility checks. We incorporate a Boundary Refinement step using \mbox{K-Nearest Neighbors (KNN)} to smooth local label inconsistencies, ensuring continuous surface semantics.}

\vspace{-4pt}
\subsection{Physics-ready geometry reconstruction}
\label{subsec:reconstruction}

\textls[+10]{Following semantic fusion, the scene is partitioned into the target object $\mathcal{P}_{\text{obj}}$ and the environmental background $\mathcal{P}_{\text{bg}}$. However, semantic identity does not guarantee geometric utility. The raw 3DGS representation contains low-density visual artefacts, including floaters, semi-transparent ghost primitives, and stretched needle-like splats, which creates false obstacles for the physics engine.}

To resolve this problem, we implement a three-stage reconstruction pipeline applied identically to both the object and background point clouds. This process systematically converts the noisy visual representation into a clean, collision-free physical environment.

\vspace{-4pt}
\subsubsection{Stage 1: intrinsic attribute filtering}
The initial phase acts as a global statistical cleaner, removing primitives that contribute to visual haze but lack physical substance. We apply two rigorous filters to the entire scene:
\begin{itemize}\setlength{\leftskip}{+0.16cm} 
    \item Opacity Threshold: We discard primitives with low opacity (e.g., $\alpha < 0.1$). This effectively eliminates the semi-transparent ``mist'' often found hovering above surfaces.
    \item Geometric Regularisation: We analyse the covariance scales to identify and remove overly stretched, ``needle-like'' primitives. These artefacts, common in sparse-view areas, are pruned to prevent the physics engine from registering false collisions with non-existent spikes.
\end{itemize}

\vspace{-4pt}
\subsubsection{Stage 2: semantic-guided connectivity pruning}
\textls[+8]{Even after statistical filtering, isolated clusters of noise (floaters) may persist. We leverage the semantic prior established in the previous section to perform topological cleaning. For any given semantic partition (whether object or background), we assume the physical entity corresponds to the dominant \mbox{geometric structure.}}

\textls[-10]{We employ DBSCAN clustering to segment the point cloud into spatially disjoint groups. By retaining only the largest connected cluster and pruning all smaller detached components, we effectively wipe out floating artefacts. This ensures that $\mathcal{P}_{\text{obj}}$ resolves to a single coherent object and $\mathcal{P}_{\text{bg}}$ resolves to a clean static environment (e.g., the table surface), free from phantom obstacles.}

\subsubsection{Stage 3: collision-ready meshing via alpha shapes}
Finally, to bridge the gap to robotic manipulation, we convert the cleaned point geometry into a mesh. We select the alpha shapes algorithm, which functions as a ``shrink-wrap'' operation. Unlike implicit smoothing methods, alpha shapes tightly conform to the point distribution, preserving sharp geometric features, such as box corners and handle edges, that are critical for stable grasping contact.

\vspace{-4pt}
\subsection{Interactive digital twin and planning}

The semantically-segmented and geometrically refined 3D models are imported into the Unity engine. The background point cloud $\mathcal{P}_{\text{bg}}$ forms the static environment. Each manipulable object in $\mathcal{P}_{\text{obj}}$ is assigned a MeshCollider for accurate collision detection and a Rigidbody for realistic physics-based interactions. The Unity environment acts as the digital twin.

\textls[+14]{We establish a seamless, high-bandwidth communication bridge between Unity and the standard \mbox{robotic software stack (ROS 2)}} \textls[-18]{using the ROS2 for Unity plug-in. This enables bidirectional state synchronisation: the robot's state is sent to Unity, and the digital twin's environment geometry and object poses (derived from the segmented point clouds) are dynamically sent to the MoveIt 2~\cite{coleman2014reducingbarrierentrycomplex}  planning scene. The system uses this complete and rapidly generated information to perform collision-aware motion planning. Generated trajectories are first validated in the physics-enabled simulation before being sent to the physical robot for execution, forming a reliable perception-to-planning-to-validation sim-to-real workflow.}

\textls[+12]{It is worth noting that the scope of this work is currently limited to static scenes. This design choice} prioritises reconstruction efficiency and geometric stability, key requirements for the proposed rapid scan-and-plan workflow, over the computational complexity associated with dynamic modelling. Furthermore, the static assumption remains valid for the targeted tabletop rearrangement tasks, where the environment is assumed to remain stable during the planning phase.

\vspace{-4pt}
\section{Experiments and results}

To comprehensively validate the effectiveness and robustness of our proposed high-fidelity 3DGS digital twin framework for robotic manipulation tasks, we designed and conducted a series of quantitative experiments. This section details the experimental setup, task definition, baselines for comparison, and quantitative analysis of key evaluation metrics.

\vspace{-4pt}
\subsection{Experimental setup}

Our experimental platform centres on a Franka Emika 7-DOF robot arm equipped with an Intel RealSense D435i RGB-D camera mounted on its end-effector. The camera provides high-resolution colour images that serve as input for our 3DGS framework. All computations were performed on a workstation equipped with an NVIDIA RTX 4090 GPU, ensuring rapid 3DGS training and rendering to meet the demanding requirements for reconstruction efficiency.

To systematically evaluate reconstruction robustness across varying geometric and optical challenges, we selected seven representative objects spanning three difficulty levels. The L1-Basic category includes a Blue Box and Yellow Cube, featuring convex geometry with Lambertian surfaces that serve as baseline objects. The L2-Complex category comprises a Toy Hammer and Scissors, presenting non-convex shapes with thin structures that challenge geometric reconstruction. The L3-Textured category contains a Diet Coke Bottle, Glue Stick, and Pen, exhibiting high-frequency surface details that test the framework's ability to capture fine visual features. These objects enable targeted analysis of reconstruction quality, semantic segmentation accuracy, and geometric fidelity across distinct challenge categories.

\begin{figure}[h]
    \centering
    \begin{subfigure}[b]{0.32\textwidth}
        \centering
        \includegraphics[width=\textwidth]{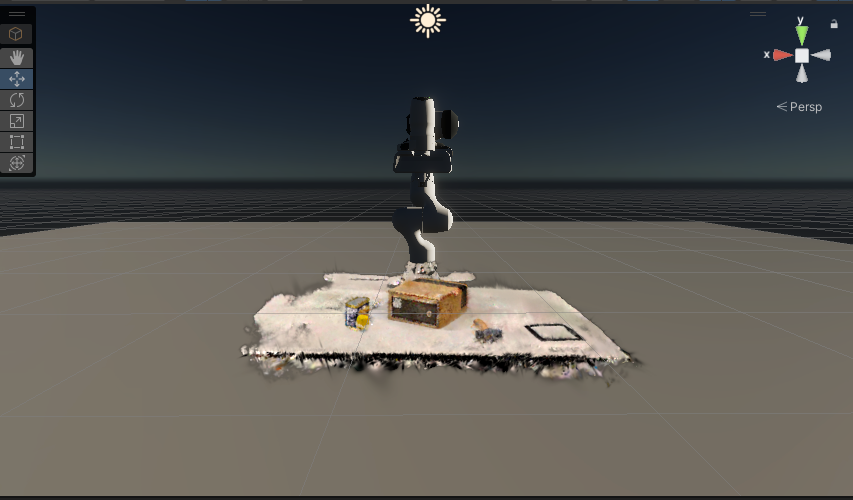}
        \caption{}
        \label{fig:unity}
    \end{subfigure}
    \hfill
    \begin{subfigure}[b]{0.32\textwidth}
        \centering
        \includegraphics[width=\textwidth]{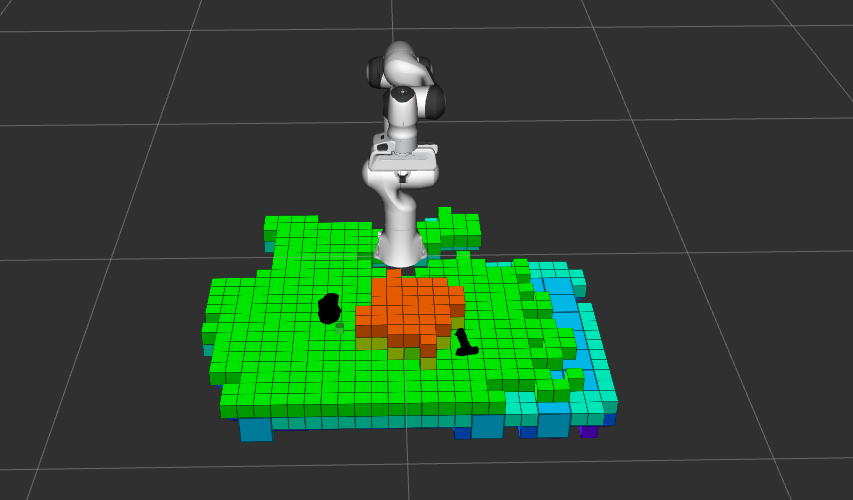}
        \caption{
        }
        \label{fig:ros}
    \end{subfigure}
    \hfill
    \begin{subfigure}[b]{0.32\textwidth}
        \centering
        \includegraphics[width=\textwidth]{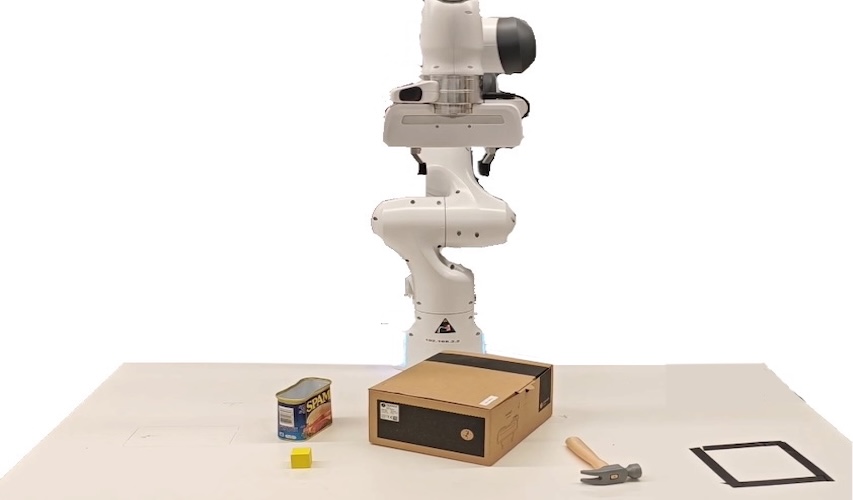}
        \caption{
        }
        \label{fig:setting}
    \end{subfigure}
    \caption{Integration and validation of the digital twin framework across simulation and reality. The Unity view (a)
shows the high-fidelity, photorealistic digital twin built with 3DGS and integrated with the physics engine. This model generates and validates collision-aware motion plans visualised in (b) the RViz interface, which uses simplified geometry for MoveIt planning. The validated plan is then executed by \mbox{(c) the real Franka Emika robot}, completing the sim-to-real workflow.}
    \label{fig:system_overview}
\end{figure}

As illustrated in Figure~\ref{fig:system_overview}, we constructed a challenging, unstructured tabletop scene to test the system's zero-shot generalisation capabilities. The scene includes objects varying in geometry, texture, and function: a Toy Hammer with complex geometric shape, a simple-surfaced Blue Box, and a small Yellow Cube. Additionally, a cardboard box serves dual roles, acting as a static obstacle initially and subsequently becoming a target placement area. This dynamic role assignment tests the system's adaptability to environmental changes.

\vspace{-18pt}
\subsubsection{Task definition}

The core evaluation task is defined as a long-horizon, zero-shot rearrangement comprising three sequential manipulation steps. First, in the object-obstacle interaction phase, the robot grasps the Blue Box and places it atop the cardboard box, testing planning and manipulation capabilities in the presence of obstacles. Second, the object-object interaction phase requires grasping the Yellow Cube and placing it on the Blue Box, demanding accurate perception and interaction with previously moved objects. Third, the irregular object manipulation phase involves grasping the geometrically complex Toy Hammer and placing it within a designated target frame, evaluating robustness in reconstructing and manipulating irregular shapes. Successful execution requires proactive planning in a dynamic environment rather than purely reactive perception-based control. Since all objects and scene layouts were unseen during system development, this constitutes a zero-shot manipulation problem.

\vspace{-4pt}
\subsubsection{Comparison methods}

To quantitatively evaluate our approach, we compared against two representative baselines focusing on 3D reconstruction fidelity and efficiency. Our 3DGS-based method performs a single scene scan using sparse multi-view RGB images (10 to 20 views). This view count was experimentally determined to represent the optimal trade-off for the “scan-and-plan'' workflow: it provides sufficient parallax for robust reconstruction while keeping the robotic data acquisition time within a practical minimum. This allows us to rapidly build a high-fidelity digital twin in minutes, which is then imported into the Unity physics engine for complete planning and pre-validation. 

Baseline 1 employs traditional point cloud reconstruction using the Intel RealSense D435i to perform multi-view depth fusion, establishing a benchmark for reconstruction efficiency and geometric accuracy. Baseline 2 utilises a state-of-the-art NeRF framework (Instant-NGP) for 3D scene reconstruction from identical sparse multi-view RGB images, enabling a direct comparison of efficiency and photorealistic quality between 3DGS and NeRF approaches in this sparse-data regime.

\vspace{-4pt}
\subsubsection{Evaluation metrics}

Our evaluation encompasses five categories of metrics. For reconstruction fidelity and efficiency, we measure reconstruction time from image capture to model completion, along with Peak Signal-to-Noise Ratio (PSNR) and Structural Similarity Index (SSIM) computed on held-out novel views. 

To assess semantic segmentation accuracy, we evaluate mean Intersection over Union (mIoU) against manually annotated ground truth, alongside two custom consistency metrics. First, we define the 3D Projection Consistency score. Let $\mathcal{V}_p$ be the set of views where a 3D point $\mathbf{p}$ projects strictly within the image sensor boundaries. We denote the number of valid observations as $N_p = |\mathcal{V}_p|$. Within these valid views, let $N_p^{\text{fg}}$ be the count of views where the projected pixel falls within the foreground mask region. The consistency score is defined as $\text{Consistency}(\mathbf{p}) = N_p^{\text{fg}} / N_p$. We report the dataset-level consistency as the percentage of valid points (where $N_p > 0$) that achieve a high confidence of $\text{Consistency}(\mathbf{p}) \geq 0.8$. Second, the Ghost Index measures artefact introduction in background regions, defined as the percentage increase in foreground points when relaxing the voting threshold, normalised by the total point count.

\textls[-10]{Geometric fidelity is evaluated using Chamfer Distance, Precision, and F1-Score against manually cleaned ground truth models. Finally, manipulation task performance is measured through task success rate, collision count, and qualitative placement correctness with respect to the designated goal regions.}

\vspace{-4pt}
\subsection{Reconstruction performance}

\textls[-12]{The quantitative results presented in Table~\ref{tab:reconstruction_performance} demonstrate the superior performance of our proposed 3DGS-based method in both efficiency and rendering quality. Our method achieves an average reconstruction }time of 229 seconds across test scenes with varying input sparsity (10 to 20 views, 3000 iterations), representing a 5$\times$ speed-up over the NeRF baseline (1123 seconds) and effectively reducing the reconstruction bottleneck from hours to minutes suitable for rapid deployment. Reconstruction time scales with scene complexity, ranging from 109 seconds for simple textured objects to 349 seconds for scenes with complex geometry or view-dependent effects.

Regarding rendering quality, our method achieves an average PSNR of 37.03 dB and SSIM of 0.9821, representing an 8 dB improvement over NeRF (28.33 dB / 0.9037). Performance varies predictably with object category: well-textured objects achieve high
 quality (PSNR 41.34 dB, SSIM 0.9934) with sparse inputs, preserving fine details such as legible text and sharp edges, while geometrically simpler objects with uniform surfaces yield intermediate performance around 36.77 dB. Although traditional point cloud approaches offer faster processing (approximately 20 seconds), they lack the photorealistic textures essential for high-fidelity digital twins and suffer from significant drift on consumer-grade depth sensors, rendering them unsuitable for our application.

\begin{table}[H]
\caption{Reconstruction performance comparison. Results averaged across test scenes with varying complexity (10 to 20 input views, 3000 iterations). Times measured on NVIDIA RTX 4090. PSNR/SSIM computed on 5 held-out novel views.}
\label{tab:reconstruction_performance}
\centering
\fontsize{10pt}{12pt}\selectfont
\begin{tabular}{lccc}
\toprule
\textbf{Method} & \textbf{Time (s) $\downarrow$} & \textbf{PSNR (dB) $\uparrow$} & \textbf{SSIM $\uparrow$} \\
\midrule
Point Cloud & $\sim$20 & N/A & N/A \\
NeRF (Instant-NGP) & 1123 $\pm$ 200 & 28.33 $\pm$ 1.0 & 0.9037 $\pm$ 0.023 \\
Ours (3DGS) & \textbf{229 $\pm$ 120} & \textbf{37.03 $\pm$ 5.0} & \textbf{0.9821 $\pm$ 0.011} \\
\bottomrule
\end{tabular}
\end{table}

\vspace{-18pt}
\subsection{Scalability and engineering cost}

To assess engineering applicability, we analyse runtime and memory usage as a function of (i) number of input views, (ii) scene size measured by the number of Gaussians after optimisation, and (iii) number of segmented objects. We summarise the observed trends below, with per-configuration measurements reported in Table~\ref{tab:scalability}.

\begin{table}[htbp!]
\centering
\fontsize{10pt}{12pt}\selectfont
\caption{Scalability of the pipeline with respect to input views and scene complexity. All times measured on an NVIDIA RTX 4090. Peak GPU memory recorded during 3DGS optimisation. \#Gaussians counts reflect scene complexity after optimisation.}
\label{tab:scalability}
\begin{tabular}{ccccc}
\toprule
\textbf{Views} & \textbf{\#Gaussians (K)} & \textbf{3DGS time (s)} & \textbf{Geo. conv. (s)} & \textbf{Peak GPU (GB)} \\
\midrule
10  & $\sim$1300  & 109 $\pm$ 18 & < 5 & 18.3 $\pm$ 0.8 \\
15  & $\sim$1700  & 229 $\pm$ 52 & < 5 & 20.1 $\pm$ 1.0 \\
20  & $\sim$2100  & 349 $\pm$ 61 & < 5 & 22.4 $\pm$ 1.2 \\
\bottomrule
\end{tabular}
\end{table}

Empirically, 3DGS optimisation time scales approximately linearly with the number of input views, ranging from approximately 109~s at 10 views to 349~s at 20 views on the RTX 4090. The geometry conversion stage (filtering, DBSCAN clustering, and alpha-shapes meshing) adds less than 10\% overhead relative to the 3DGS training time and scales primarily with the number of retained Gaussians rather than the number of input views. Semantic lifting cost grows linearly with the product of the number of views and the number of target objects, as each object requires per-view mask projection and depth clustering. Peak GPU memory during optimisation remains practical for high-end workstations; however, deployment on embedded platforms may require model pruning or lower-resolution training, as discussed in the Limitations section.

\vspace{-4pt}
\subsection{Semantic segmentation accuracy}

\vspace{+4pt}
\subsubsection{Multi-view consistency analysis}

\textls[-10]{A core challenge in lifting 2D masks to 3D is balancing completeness with noise suppression. Table~\ref{tab:multiview_consistency} presents the trade-off between consistency score and ghost index under different voting thresholds, where $N$ denotes the number of visible views. A loose threshold ($N/2.0$) achieves low artefact rate (ghost index 22.48\%) but suffers from incomplete segmentation (82.41\% consistency), often missing object boundaries. Conversely, a strict threshold ($N/1.0$) guarantees 100\% consistency but introduces excessive noise (ghost index 67.23\%), generating floating obstacles that interfere with motion planning. We selected $N/1.5$ as the operating point, achieving 93.72\% consistency while maintaining the ghost index below 50\%, ensuring robust 3D object definition without compromising the free space required for collision-free planning.}

\begin{table}[H]
\caption{Multi-view consistency versus artefact rate at different voting thresholds. Results measured on 7-object benchmark with 15 views per scene.}
\label{tab:multiview_consistency}
\centering
\fontsize{10pt}{12pt}\selectfont
\begin{tabular}{lcc}
\toprule
\textbf{Voting Threshold} & \textbf{Consistency (\%) $\uparrow$} & \textbf{Ghost Index (\%) $\downarrow$} \\
\midrule
$N/2.0$ & 82.41 & 22.48 \\
$N/1.8$ & 87.69 & 37.42 \\
$N/1.5$ & \textbf{93.72} & \textbf{46.27} \\
$N/1.2$ & 100.0 & 53.19 \\
$N/1.0$ & 100.0 & 67.23 \\
\bottomrule
\end{tabular}
\end{table}

\vspace{-18pt}
\subsubsection{Overall semantic quality}

To assess overall semantic understanding, we computed the mean Intersection over Union (mIoU) between the projected semantic masks and manually annotated foreground masks. Our framework achieves a 2D segmentation mIoU of 0.87 averaged across all views and objects, and 3D projection consistency reaches 0.93. These results confirm that our multi-view fusion approach effectively bridges 2D perception and 3D geometric reconstruction, providing a reliable semantic layer for robotic manipulation.

\subsection{Ablation study on point cloud cleaning}

\begin{figure}[H]
    \centering
    \includegraphics[width=0.8\textwidth]{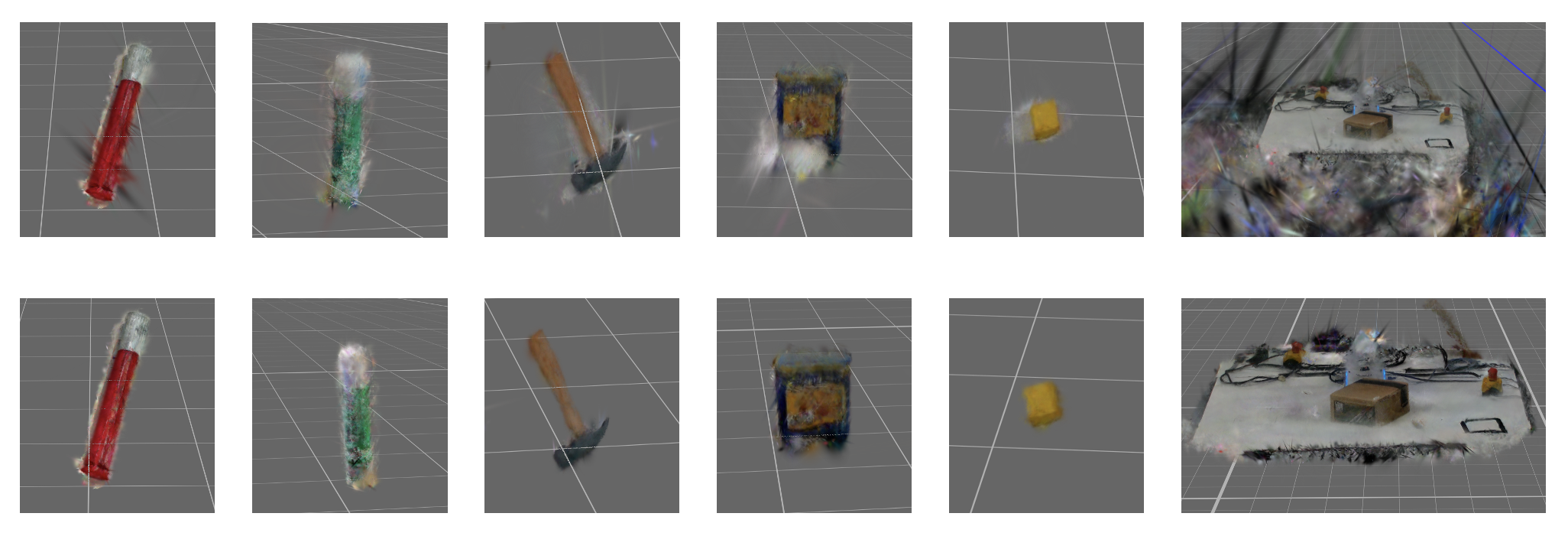}
    \vspace{-3mm}
    \caption{Qualitative results of the point cloud cleaning pipeline. Top: Raw 3DGS point clouds exhibit floaters and surface fuzziness, which impede precise collision checking. Bottom: Refined geometries after applying our multi-stage filtering, including heuristic filtering and DBSCAN. The process effectively removes artefacts and sharpens boundaries, yielding planning-ready digital twins for manipulation tasks.}
    \label{fig:point_cloud_cl}
\end{figure}

To validate the effectiveness of each component within our cleaning pipeline, we conducted a rigorous ablation study on four representative objects from the L1-Basic and L2-Complex categories (Blue Box, Yellow Cube, Toy Hammer, Scissors). The raw point cloud generated by 3DGS typically contains floaters and ghosting artefacts that compromise geometric accuracy. To qualitatively illustrate this issue and the effectiveness of our cleaning pipeline, we present representative results in Figure~\ref{fig:point_cloud_cl}. Raw 3DGS reconstructions contain floating artefacts and surface fuzziness, which can create unreliable geometry for collision checking and motion planning. After applying the proposed multi-stage filtering strategy, the refined point clouds show clearer object boundaries and improved structural consistency. This qualitative comparison shows why geometric cleaning is required before converting 3DGS reconstructions into planning-ready digital twins.

We compared four configurations: the original 3DGS output serving as baseline; denoising only, applying attribute-based filtering (opacity and colour thresholds); clustering only, applying DBSCAN spatial clustering ($\text{eps}=0.02$, $\text{min\_samples}=10$); and our full method combining both components sequentially. Results are presented in Table~\ref{tab:geometry_results_ablation}.

\begin{table}[H]
\centering
\fontsize{10pt}{10pt}\selectfont
\caption{\textls[-12]{Ablation study on geometric fidelity of the cleaning pipeline. Results averaged across four test objects. Ground truth obtained via manual cleaning in CloudCompare. Chamfer Distance computed at 1mm resolution.}}
\label{tab:geometry_results_ablation}
\begin{tabular}{lccc}
\toprule
\textbf{Method} & \textbf{Chamfer Distance
$\downarrow$} & \textbf{Precision $\uparrow$} & \textbf{F1-Score $\uparrow$} \\
\midrule
Original & 0.0052 & 0.8846 & 0.9369 \\
Denoising Only & 0.0055 & 0.8838 & 0.9369 \\
Clustering Only & 0.0043 & 0.8958 & 0.9429 \\
Denoising + Clustering & \textbf{0.0020} & \textbf{0.9977} & \textbf{0.9989} \\
\bottomrule
\end{tabular}
\end{table}

\vspace{-4pt}
\textls[+5]{The results reveal the contribution of each component. Applying clustering alone provides noticeable improvement over baseline, reducing Chamfer distance from 0.0052 to 0.0043 and increasing F1-Score from 0.9369 to 0.9429, indicating effectiveness in removing spatially isolated noise. Interestingly, applying de-noising in isolation yields no improvement (Chamfer distance slightly increased to 0.0055),} suggesting that attribute-based filtering alone cannot handle complex artefacts where floaters remain spatially connected to the main body. However, the full method achieves substantial improvements, reducing Chamfer distance to 0.0020 and elevating F1-Score to 0.9989. This demonstrates a synergistic effect: attribute-based filtering first removes low-confidence primitives, while clustering then removes spatially isolated components that remain after filtering. This two-stage process is essential for producing high-fidelity point clouds required for reliable manipulation.

\vspace{-4pt}
\subsection{Real-world robotic validation}

\begin{figure}[H]
\centering
    \begin{subfigure}[b]{\textwidth}
        \centering
        \includegraphics[width=\textwidth]{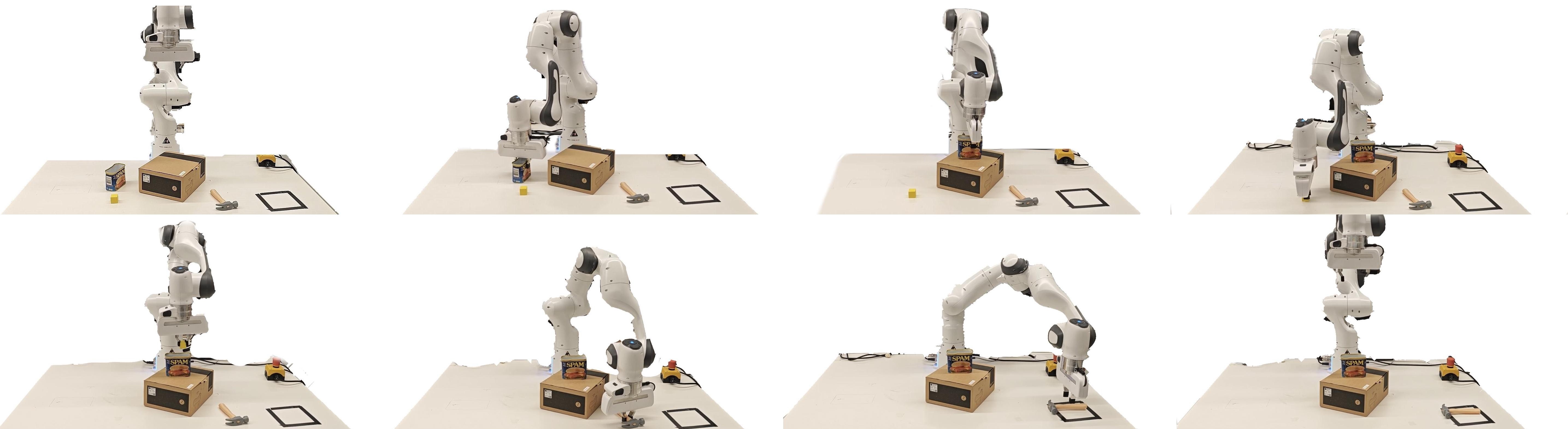}
        \caption{
        }
        \label{fig:realbotaction}
    \end{subfigure}
    \hfill
    \begin{subfigure}[b]{\textwidth}
        \centering
        \includegraphics[width=\textwidth]{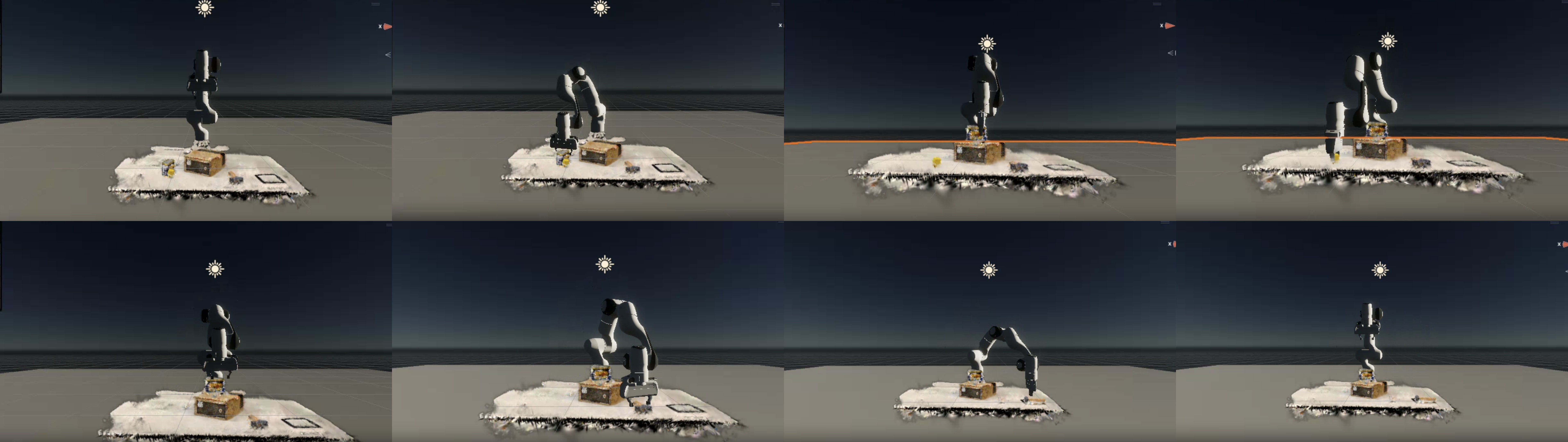}
        \caption{
        }
        \label{fig:sim}
    \end{subfigure}
    \caption{\textls[-8]{Execution sequence of the multi-step rearrangement task in (a) the real world and (b) the digital twin. The robot grasps the Blue Box and places it on the cardboard box, then grasps the Yellow Cube and stacks it on the Blue Box, and finally grasps the Toy Hammer and places it in the target area. This demonstrates the framework's capability for complex, zero-shot manipulation with proactive planning validated in simulation.}}
    \label{fig:manipulation_sequence}
\end{figure}

We finally validate this framework on a Franka Emika arm to demonstrate its ability to enable successful real-world manipulation. We conducted 10 independent trials of the long-horizon rearrangement task, with the complete execution sequence visualised in Figure~\ref{fig:manipulation_sequence}.

Success criteria:
To ensure rigorous evaluation, we define a trial as successful only if it meets three conditions: (1) the robot successfully detects and grasps the correct target object; (2) the object is transported and placed stably within the designated goal region; and (3) the entire trajectory is collision-free with respect to both static obstacles and the environment.

\textls[-8]{Results analysis:
Under these strict criteria, our framework achieved a 100\% success rate in simulation validation and a 90\% success rate in real-world execution (9/10 trials). The solitary failure occurred during the grasping attempt of the 2.5 cm Yellow Cube. Due to the object's diminutive scale, a minor gripper alignment error resulted in a missed grasp. This failure case highlights the high-precision challenges inherent in manipulating small-scale objects that approach the resolution limits of the 3DGS-based geometric reconstruction and gripper finger geometry.}

In terms of placement accuracy, qualitative assessment confirmed that all manipulated objects were correctly deposited strictly within the designated regions. While exact metric error was not instrumented, this consistent alignment demonstrates that the system effectively satisfied the spatial tolerances required for the rearrangement task. Importantly, zero collisions were observed during any trial, validating the high fidelity of our collision geometry generation. These results demonstrate that 3DGS-based digital twins, combined with semantic and geometric consistency, provide a reliable foundation for complex manipulation in unstructured environments.

\vspace{-18pt}
\section{Limitations}

Despite the demonstrated effectiveness of our framework, several limitations warrant acknowledgment to set realistic expectations for deployment and guide future improvements.

Static-scene assumption: Our current system performs a one-time reconstruction and assumes objects remain stationary during the planning and execution phases (typically 5--15 seconds in our experiments). This design choice prioritises reconstruction efficiency and geometric stability, which are key requirements for the proposed rapid scan-and-plan workflow. However, it limits applicability to dynamic scenarios involving deformable objects, moving humans, or objects that shift during task execution. The framework is specifically designed for tabletop rearrangement tasks where the environment remains stable during planning, a common assumption in structured manipulation settings. Extending to dynamic scenes would require integrating continuous reconstruction methods or dynamic 3DGS variants, as discussed in Section~6.

\textls[-12]{Limited view-count validation: Our experiments focus on sparse-view reconstruction using \mbox{10--20 input} views,} which represents the optimal trade-off between reconstruction quality and data acquisition time for desktop-scale manipulation (approximately 2--3 minutes of robotic scanning). While this range is sufficient for our target scenarios, we have not systematically validated performance with significantly higher view counts (e.g., 30--50 views) or explored the potential quality improvements from denser sampling. The scaling behaviour beyond 20 views, including potential diminishing returns in reconstruction fidelity versus increased computational cost, remains an open empirical question.

\textls[-12]{Scene scale constraints: The framework is optimised for desktop-scale manipulation scenarios with workspace dimensions of approximately 0.3--1.0 m depth and objects ranging from 5--30 cm in size. Our DBSCAN clustering parameters (eps = 0.02m, min\_samples = 10) and semantic fusion thresholds are calibrated for this scale.} Deployment in substantially different environments—such as warehouse-scale logistics (multi-meter workspaces), micro-manipulation (< 1 cm objects), or outdoor unstructured settings—would require recalibration of spatial thresholds and potentially architectural modifications to handle the increased scene complexity and point cloud density.

Object geometry constraints: The current pipeline is optimised for rigid, opaque objects with diffuse or mildly specular surfaces, as evidenced by our test objects (boxes, tools, textured bottles). Highly challenging cases such as transparent objects (glass containers), highly reflective surfaces (polished metal), thin wire-like structures (< 5 mm diameter), or materials with complex subsurface scattering may produce incomplete 3DGS reconstructions or inaccurate collision geometry. These failure modes stem from fundamental limitations in multi-view RGB-based reconstruction rather than our specific processing pipeline, but they constrain the generality of the approach.

\vspace{-4pt}
\section{Conclusion and future direction}

\textls[+8]{This paper presented a holistic, closed-loop framework for rapidly creating high-fidelity, interactive digital twins for robotic manipulation from sparse RGB views. The approach combines 3DGS~\cite{kerbl20233dgaussiansplattingrealtime} for fast photorealistic reconstruction, Grounded SAM~\cite{ren2024groundedsamassemblingopenworld} for zero-shot semantic segmentation, and a filtering pipeline to generate clean, planning-ready collision geometry. The system completes reconstruction in under 4 minutes on average (229~s) while achieving high visual fidelity (37.03~dB PSNR and 0.9821 SSIM), representing a 5-fold speed-up over the NeRF-based baseline.}

Real-world experiments on a long-horizon rearrangement task demonstrate the practical utility of the framework. Motion plans generated and validated within the digital twin achieved a 90\% success rate when executed on a Franka Emika robot, with all successful placements falling within the predefined task regions. These results provide evidence that the framework effectively addresses the sim-to-real gap, enabling reliable robot operation in unstructured environments.

The multi-stage de-noising and meshing pipeline proved essential for converting the unstructured 3DGS output into planner-compatible geometries. The ablation study confirms that both heuristic filtering and cluster-based de-noising contribute synergistically, achieving a near-perfect F1-score of 0.9989 against manually cleaned ground truth models. This Gaussian-to-mesh conversion represents a critical bridge between modern neural rendering and traditional motion planning frameworks.

Several directions warrant future investigation. First, the current framework assumes static scenes and performs a one-time reconstruction. Integrating dynamic 3DGS variants \cite{zhou2023drivinggaussian} or implementing continuous update mechanisms would enable the digital twin to remain consistent with evolving environments. Second, the system currently models only geometry and appearance. Incorporating methods for online physical property estimation \cite{cherian2024llmphycomplexphysicalreasoning} would enable more sophisticated manipulation strategies involving contact-rich interactions. Third, while this work focuses on motion planning given predefined grasps, integrating robust grasp planning modules \cite{ji2024graspsplats} that operate directly on Gaussian representations would further automate the pipeline.

A particularly promising direction involves leveraging the digital twin as an enabler for learned policies. The framework could serve dual purposes: as a safety validation platform where policies are tested through thousands of simulated iterations before deployment, and as a data generation engine that autonomously creates diverse training datasets without physical resource consumption. This capability could significantly accelerate the development of vision-language-action models and reinforcement learning approaches for manipulation.

The work demonstrates that the synergy between 3DGS efficiency, foundation model capabilities, and robust geometric processing provides a practical paradigm for robotic manipulation in unstructured environments. By unifying perception, reconstruction, and planning into a closed-loop system, the framework represents a step toward autonomous robots that can rapidly adapt to novel surroundings with both speed and reliability.

\section*{Data availability statement}

To support reproducibility, all code, datasets, and documentation are available at: \url{https://github.com/535A59/3DGS-Digital-Twin}.

\section*{Declaration of Generative AI and AI-assisted Technologies}

During the preparation of this manuscript, the authors used ChatGPT only to improve language and readability in limited sections of the manuscript. This tool was not used for scientific content generation, data analysis, result interpretation, figure generation, or idea development. After using this tool, the authors reviewed and edited the content as needed and take full responsibility for the content of this manuscript.

\section*{Acknowledgments}
\textls[-18]{This work was partially supported by the Advanced Research and Invention Agency [grant number SMRB-SE01-P06] and NVIDIA Academic Grant Program.}

\section*{Authors' contribution}
\textls[-18]{Ziyang Sun: conceptualisation, methodology, software, validation, formal analysis, investigation, data curation, visualisation and writing---original draft; Lingfan Bao: methodology, validation, and writing---review and editing; Tianhu Peng: writing---review and editing; Jingcheng Sun: writing---review and editing. Chengxu Zhou: conceptualisation, resources, supervision, project administration, funding acquisition and writing---review and editing. All authors have read and agreed to the published version of the manuscript.}

\section*{Conflicts of interest}
The authors declare no competing interests.

\bibliographystyle{ieeetr}
\bibliography{ref}

\end{document}